\documentclass[11pt]{article}
\usepackage{graphicx}        
\usepackage{amssymb}
\usepackage[numbers]{natbib}
\usepackage{bm}        
\usepackage{tikz}
\usepackage{smartdiagram}
\usepackage{algorithm2e}

\newcommand{\ff}[1]{{{\boldsymbol{#1}}}} 
\def\x{{\boldsymbol{x}}}

\def\be{\begin{equation}}
\def\ee{\end{equation}}

\begin{document}

\title{Nature-Inspired Algorithms in Optimization: Introduction, Hybridization and Insights}
\author{Xin-She Yang \\
School of Science and Technology, \\ Middlesex University London, \\ 
The Burroughs, London NW4 4BT, United Kingdom.  }

\date{}

\maketitle

\abstract{Many problems in science and engineering are optimization problems, which may require sophisticated optimization techniques to solve. Nature-inspired algorithms are a class of metaheuristic algorithms for optimization, and some algorithms or variants are often developed by hybridization. Benchmarking is also important in evaluating the performance of optimization algorithms. This chapter focuses on the overview of optimization, nature-inspired algorithms and the role of hybridization. We will also highlight some issues with hybridization of algorithms.}

\bigskip 

{\bf Keywords:} 
Algorithm, Benchmark, Hybrid Algorithms, Nature-Inspired Algorithms, Optimization.

\bigskip
\noindent {\bf Citation Details:}
Xin-She Yang, Nature-Inspired Algorithms in Optimization: Introduction, Hybridization, and Insights,  in: {\it Benchmarks and Hybrid Algorithms in Optimization and Applications} (Edited by Xin-She Yang), Springer Tracts in Nature-Inspired Computing, pp. 1--17 (2023). \\
https://doi.org/10.1007/978-981-99-3970-1\_1
\\[15pt]

\section{Introduction}

Nature-inspired algorithms and their various hybrid variants have become popular in recent years for solving optimization problems,
due to their flexibility and stable performance. In many applications
related to science and engineering, problems can often be formulated as optimization problems with design objectives, subject to various constraints. A typical optimization problem consists of one objective, subject to various inequality and inequality constraints. The objectives can be the main goal to be optimized, such as the minimization of cost, energy consumption, travel distance, travel time, CO$_2$ emission, wastage, and environmental impact, or the maximization of efficiency, accuracy, and performance as well as
sustainability.

Almost all design problems have design constraints or requirements.
Constraints can be design requirements, such as physical dimensions, capacity, budget, design codes/regulation, time and other quantities such as stress and strain requirements. They are often written as mathematical inequalities or equalities. Due to the nonlinear nature of such optimization problems, sophisticated optimization algorithms and techniques are required to solve them. In the current literature, there are many optimization techniques and algorithms for solving optimization problems.

In the last few decades, gradient-free nature-inspired algorithms have received a lot of attention with significant developments. One class of such nature-inspired algorithms are based on swarm intelligence~
\citep{Kennedy1995PSO,YangBook2014,YangHe2019,Yang2020Rev}. The literature of nature-inspired algorithms and swarm intelligence is expanding rapidly, here we will introduce some of the most recent and widely used nature-inspired optimization algorithms.

\section{Optimization and Algorithms}

Before we introduce some nature-inspired algorithms in detail, let us discuss briefly the four key components of optimization and their related issues.

\subsection{Components of Optimization}

\begin{figure}[h]
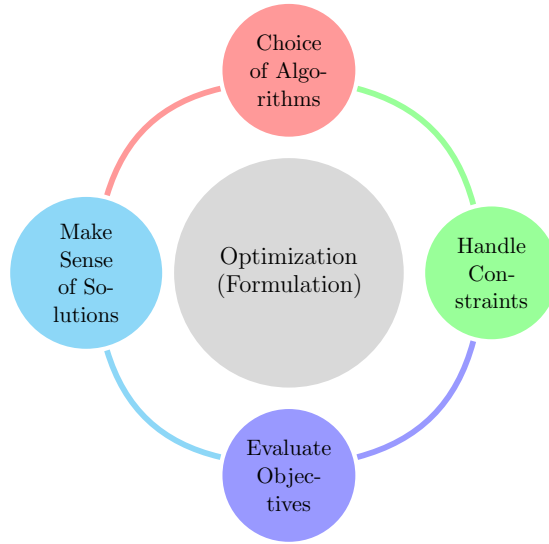

\begin{center}
\smartdiagramset{planet text width=2.5cm}
\scalebox{0.77}{
\smartdiagram[connected constellation diagram]{Optimization \\ (Formulation), Choice of Algorithms, Make Sense of Solutions, Evaluate Objectives, Handle Constraints}
}
\caption{Important components of optimization.}
\end{center}
\end{figure}

Once an optimization problem has been formulated properly with the right objective and the correct constraints, the next steps will be to find the optimal solutions using an efficient algorithm or optimization technique. In general, to solve an optimization problem involves four main components: The choice of algorithm, handling the constraints, evaluation of the objective function, and making sense of the solutions.

\begin{itemize}
\item \emph{Choice of algorithms}: To solve any optimization problem, an efficient algorithm, or a sufficiently good algorithm, should be selected. In many cases, the choice may not be easy, because either there are many different algorithms to choose from or there may not be any efficient algorithms at all. In many cases, the choice of algorithm may depend on the type of problem, expertise of the user, the availability of the computational resource, the time constraint, quality of the desired solutions and other factors.

\item \emph{Handling the constraints}: Even an efficient algorithm is used for solving an optimization problem, the handling of constraints is an important part of problem solving. Otherwise, the solutions obtained may not satisfy all the constraints, leading to infeasible solutions. There are many constraint-handling techniques, such as the penalty method, dynamic penalty, evolutionary method, epsilon-constraint method, and others. A good choice of proper constraint-handling techniques will help to ensure the solution quality.

\item \emph{Evaluation of the objective}: Depending on the type of optimization problems, the evaluation of the objective functions can be a very time-consuming part. For function optimization, such evaluations are straightforward. However, for many design problems such as protein folding and aerodynamic design problems, each evaluation of such objective values can take hours or even days due to the extensive use of external simulators or software packages. In any good optimization procedure, the number of objective evaluations should be minimized so as to save time and cost.

\item \emph{Make sense of the solutions}: Once a feasible set of solutions are obtained, users or designers have to make sense of the solutions by checking if all constraints are satisfied, understanding what these solutions may imply, figuring out the stability and robustness of the solutions and then deciding which solution(s) to use for the design and further refinement.
    For single-objective optimization problems, this may not cause any issues in selecting an optimal solution. However, for multi-objective optimization problems, multiple options from the Pareto front  will be available, the choice may require some higher-level criteria or decision-makers to make the final choice, by considering other factors that may not be implemented in the optimization problems.

\end{itemize}

Due to the complexity of real-world optimization problems, it is usually challenging to obtain satisfactory results, while maintaining all the relevant interacting components to be suitable for solving the optimization problem under consideration. For the rest of this chapter, we will focus on algorithms.

\subsection{Gradients and optimization}

Traditional optimization techniques, such as Newton-Raphson based methods, use first-order derivatives or gradients to guide the search. From a mathematical perspective, if the objective is sufficiently smooth, the optimal solutions should occur at critical points where $f'(x)=0$ or at the boundaries. In this case, gradients provide the
key information needed for finding the locations of the possible optima.

Even for smooth objectives without any constraints, it can become complicated when $f(x)$ is highly nonlinear with multiple optima.
One well-known example is to find the maximum value of $f(x)=\textrm{sinc}(x)=\sin(x)/x$ in the real domain. If we can naively use
\be f'(x)=\Big[\frac{\sin(x)}{x}\Big]'=\frac{x \cos(x)-\sin(x)}{x^2}=0,   \ee
we have an infinite number of solutions for $x \ne 0$. There is no simple formula for these solutions, thus a numerical method has to be used to calculate these solutions. Even with all the efforts to find these solutions (it may not be easy in practice), we have to be careful because the true global maximum $f_{\max}=1$ occurs at $x_*=0$. This highlights the potential difficulty for nonlinear, multimodal problems with multiple optima.

Obviously, the requirement for smoothness may not be satisfied at all. For example, if we try to find the optimal solution by using $f'(x)=0$ for
\be f(x)=|x| \exp[-\sin(x^2)], \ee
we will not be able to use this condition because $f(x)$ is not differentiable at $x=0$, but the global minimum $f_{\min}=0$ indeed occurs at $x_*=0$. This also highlights an issue that optimization techniques that require the calculation of derivatives will not work for non-smooth objective functions.
High-dimensional problems can become more challenging. For example, the nonlinear function \citep{YangBook2014}
\be f(\x)=\Big \{ \Big[\sum_{i=1}^n \sin^2(x_i) \Big]
-\exp\big(-\sum_{i=1}^n x_i^2 \big) \Big\} \cdot \exp \Big[-\sum_{i=1}^n \sin^2 \sqrt{|x_i|} \Big], \ee
where $-10 \le x_i \le 10$ (for $i=1,2,...,n$), has the global minimum $f_{\min}=-1$ at $\x_*=(0,0, ...,0)$, but this function is not differentiable at the optimal point $\x_*$.

Therefore, in order to solve different types of optimization problems, we have to have a variety of optimization techniques so that they can use gradient information when appropriate, and do not use it when it is not well defined or not easily calculated. In addition, constraints, especially nonlinear constraints, tend to make the search domain irregular and even potentially with isolated regions. This will make such problems more challenging to solve. To complicate things further, we may have several objective functions instead of just one function for some design problems, and multiple Pareto-optimal solutions are sought. This will in turn make it more challenging to solve.

\section{Nature-Inspired Algorithms}

A diverse range of nature-inspired algorithms and their applications can be found in the recent literature \citep{YangBook2014,Yang2013CSFA,YangPapa2016,Yang2018Wiley}.
Now we will briefly introduce some of the most recent nature-inspired algorithms.

\subsection{Recent Nature-Inspired Algorithms}

Our intention here is not to list all the algorithms, which is not possible. Instead, we would like to use a few algorithms as examples to highlight the main components and mechanisms that can be used to carry out effective optimization in the solution or search space.

\subsubsection{Particle Swarm Optimization}
\index{particle swarm optimization} \index{PSO}

Particle swarm optimization (PSO), developed by Kennedy and Eberhart in 1995, intends  to simulate the swarming characteristics of birds and fish \citep{Kennedy1995PSO}. For the simplicity of discussions, we now use the following notations: $\x_i$ and $\ff{v}_i$ denote the position (solution) and velocity, respectively,  of a particle or agent $i$, for a population of $n$ particles, thus we have $i=1,2,...,n$.

Both the position of a particle $i$ and its velocity are iteratively updated by
\be \ff{v}_i^{t+1}= \ff{v}_i^t  + \alpha \ff{\epsilon}_1
[\ff{g}^*-\x_i^t] + \beta \ff{\epsilon}_2 [\x_i^*-\x_i^t],
\label{pso-speed-100}
\ee
\be \x_i^{t+1}=\x_i^t+\ff{v}_i^{t+1}, \label{pso-speed-140} \ee
where $\ff{\epsilon}_1$ and $\ff{\epsilon}_2$ are two uniformly distributed random numbers in [0,1]. The parameters $\alpha$ and $\beta$ are usually in the range of [0,2]. Here, $\ff{g}^*$ is the best solution found so far by all the particles in the population, often considered as some sort of centre of the swarm (not the actual geometrical centre). In addition, each individual particle has its own individual best solution $\x_i^*$ during its iteration history.

There are thousands of articles about PSO with a diverse range of applications \citep{Kennedy1995PSO,Engel2005}. However, there are some drawbacks because PSO can often have so-called premature convergence when the population loses diversity and thus gets stuck locally.
Various improvements and modifications have been developed in recent years with more than two dozen different variants. Their performance varies with different degrees of improvements.

One simple and yet quite efficient variant is the accelerated particle swarm optimization (APSO), developed by Xin-She Yang in 2008 \citep{Yang2008book}. APSO does not use velocity, but only use
the position or solution vector, which is updated in a single step
\be  \x_i^{t+1}=(1-\beta) \x_i^t+\beta \ff{g}^* +\alpha \ff{\epsilon}_t,  \ee
where $\alpha$ is a scaling factor that controls the randomness.
The typical values for this accelerated PSO are $\alpha \approx 0.1 \sim 0.4$ and $\beta \approx 0.1 \sim 0.7$. Here, $\ff{\epsilon}_t$ is a vector of random numbers, drawn from a normal distribution.
In order to reduce the randomness as the iterations continue,
a further modification and improvement to the accelerated PSO is to
use a monotonically decreasing function such as
\be \alpha=\alpha_0 \gamma^t, \quad  (0<\gamma<1), \ee
where $t$ is a pseudo-time or iteration counter. The initial value of $\alpha_0=1$ can be used for most cases.

\subsubsection{Bat Algorithm}

Based on the echolocation characteristics of microbats,  the bat algorithm (BA), developed by Xin-She Yang in 2010, uses some frequency-tuning $f$ and variations of pulse emission rate $r$ and
loudness $A$ \cite{YangBat2010,Yang2011MOBA} to update the position vectors in the search space. For bat $i$ with position $\x_i$ and velocity $\ff{v}_i$, the updates are carried out by
\be f_i =f_{\min} + (f_{\max}-f_{\min}) \beta, \label{f-equ-150} \ee
\be \ff{v}_i^{t} = \ff{v}_i^{t-1} +  (\x_i^{t-1} - \x_*) f_i , \ee
\be \x_i^{t}=\x_i^{t-1} + \ff{v}_i^t,  \label{f-equ-250} \ee
where $\beta \in [0,1]$ is a random vector drawn from a uniform distribution so that the frequency can vary from $f_{\min}$ to $f_{\max}$. In the above equations, $\x_*$ is the best solution found so far by all the virtual bats up to the current iteration $t$.

In the BA, the effective control of exploration and exploitation is achieved  by varying  loudness $A(t)$ from a high value to a lower value and simultaneously varying the emission rate $r$ from a lower value to a higher value. Mathematically speaking, the variations take the form of
\be A_i^{t+1}=\alpha A_i^t, \quad r_i^{t+1}= r_i^0 (1-e^{-\gamma t}), \quad 0<\alpha<1, \quad \gamma>0. \ee
Numerical simulation shows that BA can have a faster convergence rate
in comparison with PSO. Various studies have extended the BA to solve  multiobjective optimization with various variants versions and  applications  \citep{Yang2011MOBA,YangGandomi2012BA,Bekdas2018,Osaba2016BA,Osaba2019BA,Jayab2018Bat}. A recent study also proved its global convergence \citep{Chen2018}.

\subsubsection{Firefly Algorithm}

The firefly algorithm (BA) was developed  by Xin-She Yang developed in 2008 \citep{Yang2008book,Yang2009FA}, inspired by the light-flashing behaviour of tropical fireflies. FA uses the position vector $\x_i$ for firefly $i$ and its brightness to associate with the fitness or landscape of the objective. The solution or position is then updated
iteratively by
\be  \x_i^{t+1} =\x_i^t + \beta_0 e^{-\gamma r^2_{ij} }
(\x_j^t-\x_i^t) + \alpha \; \ff{\epsilon}_i^t, \label{FA-equ-100} \ee
where $\beta_0$ is the attractiveness parameter, and $\alpha$ is a scaling factor controlling the step sizes. Parameter $\gamma$ can be considered as a tunable parameter to control the visibility of the fireflies (and thus search modes). Here, $r_{ij}$ represents the distance between firefly $i$ at $\x_i$ and firefly $j$ at $\x_j$.

During each iteration, a pair comparison is carried out for evaluating the relative fitness among all fireflies. Briefly speaking, all the main steps of FA can be outlined as the pseudocode in Algorithm~\ref{Alg-1}.

\begin{algorithm}
\caption{Firefly algorithm. \label{Alg-1}}
\hrule
Initialize all the parameters $\alpha, \beta, \gamma$, and
population size $n$\;
Determine the light intensity/fitness at $\x_i$ by $f(\x_i)$\;
\While{$t<$ MaxGeneration}{
\For{All fireflies ($i=1:n$)}
{\For{All other fireflies ($j=1:n$) with $i \ne j$ (inner loop)}{
\If{Firefly $j$ is better/brighter than $i$}
{Move firefly $i$ towards $j$ using Eq.(\ref{FA-equ-100})\;}}
Evaluate each new solution\;
Accept the new solution if better\; }
Rank and update the best solution found\;
Update iteration counter $t \gets t+1$\;
Reduce $\alpha$ (randomness strength) by a factor $0<\delta<1$\;
} \hrule
\end{algorithm}

The role of $\alpha$ is subtle, controlling the strength of the randomness or perturbation term in the FA. In principle, randomness should be gradually reduced so as to speed up the overall convergence. For example, we can use
\be \alpha=\alpha_0 \delta^t, \ee
where $\alpha_0$ is the initial value and $0<\delta<1$ is a reduction factor. Parametric studies show that $\delta=0.9$ to $0.99$ can be used in most cases.

By analyzing characteristics of different algorithms, we can highlight some significant differences between FA and PSO. Mathematically speaking, FA is a nonlinear system, whereas PSO is a linear system. Numerical experiments have shown that FA has an ability of multi-swarming, but PSO cannot. In addition, PSO uses velocities and thus has some drawbacks. In contrast, FA does not use any velocities. Most importantly, nonlinearity in FA enriches the search behaviour and thus makes it more effective in dealing with multi-modal optimization problems~\citep{YangBook2014,Fister2013FA,Yang2013CSFA,Yang2018SwarmRev,Yang2020Swarm}.
A simple Matlab code of the standard
firefly algorithm  can be found at the Mathworks website\footnote{http://www.mathworks.co.uk/matlabcentral/fileexchange/29693-firefly-algorithm}.

\subsubsection{Cuckoo Search}

Cuckoo search (CS) algorithm is another nature-inspired optimization algorithm. CS was developed by Xin-She Yang and Suash Deb
in 2009 \citep{YangDeb2009CS,Yang2013MOCS,Yang2013CSFA},
inspired by the brood parasitism of some cuckoo species and interactions between cuckoo-host species.

The position vectors in the CS are  updated iteratively in two different ways: local search and global search with a switch probability $p_a$. The local search is carried out by
\be \x_i^{t+1}=\x_i^t +\alpha s \otimes H(p_a-\epsilon) \otimes (\x_j^t-\x_k^t), \label{CS-equ1} \ee
where $s$ is the step size, and $\x_j^t$ and $\x_k^t$ are two different solutions that are randomly selected by random permutation. Here, the Heaviside function
$H(u)$ is controlled by the switch probability $p_a$ and a random number $\epsilon$, drawn from a uniform distribution.

The global search is carried out via L\'evy flights by
\be \x_i^{t+1}=\x_i^t+\alpha L(s,\lambda), \label{CS-equ2} \ee
where the step size $s$ is drawn from a L\'evy distribution that can be approximated by a power-law distribution with a long tail
\be L(s,\lambda) \sim \frac{\lambda \Gamma(\lambda) \sin (\pi \lambda/2)}{\pi}
\frac{1}{s^{1+\lambda}}, \quad (s \gg 0). \label{Levy-eq-100} \ee
Here $\alpha>0$ is the step size scaling factor.

Various studies have shown that CS can be very efficient in finding
global optimality in many applications\cite{Yang2013MOCS,YangBook2014}.

\subsubsection{Flower Pollination Algorithm}

Though  the flower pollination algorithm (FPA), developed by Xin-She Yang and his collaborators, is not a swarm intelligence based algorithm, it is a population-based, nature-inspired algorithm. FPA was developed, inspired by the pollination characteristics of flowering plants \cite{Yang2012FPA,YangBook2014}, mimicking
the characteristics of biotic and abiotic pollination as well as co-evolutionary flower constancy.

The update of the solution vectors are realized by both local and global pollination characteristics search. They are
\be \x_i^{t+1}=\x_i^t + \gamma L(\lambda) (\ff{g}_* - \x_i^t), \ee
\be \x_i^{t+1} =\x_i^t + U (\x_j^t - \x_k^t), \ee
where $\ff{g}_*$ is the best solution vector found so far. Here, $\gamma$ is a scaling parameter, $L(\lambda)$ is a vector of random numbers, drawn from a L\'evy distribution governed by the exponent $\lambda$, in the same form given in (\ref{Levy-eq-100}).
In addition, $U$ is a uniformly distributed random number.

FPA has been applied to solve many optimization problems such as multi-objective optimization, photovoltaic parameter estimation, economic and emission dispatch, and EEG-based identification \citep{Yang2012FPA,Alam2015FPA,Abdel2019,Bekdas2015FPA,Rodrigues2016,Aly2018FPA}.
A demo Matlab code of the basic flower pollination algorithm  can be downloaded from the Mathworks website\footnote{http://www.mathworks.co.uk/matlabcentral/fileexchange/45112}.

\subsection{Other Nature-Inspired Algorithms}

In recent years, many other algorithms have appeared. An incomplete survey suggests that more than 200 nature-inspired algorithms and variants have been published in the recent literature~\citep{YangBook2014,Das2011,Aly2018FPA,Abdel2019,Yang2020Swarm}.
Obviously, it is not possible to list all the variants and algorithms.
For simplicity and for the purpose of diversity, we now list a selection of swarm intelligence (SI) based algorithms and other metaheuristic algorithms.  Examples of other swarm intelligence based algorithms are:

\begin{itemize}
\item Ant colony optimization~\citep{Dorigo1992}
\item Artificial bee colony~\citep{Karaboga2005,Karaboga2008}
\item Bees algorithm~\citep{Pham2005,Yang2005Bee}
\item Dolphin echolocation~\citep{Kaveh2013Dolphin}
\item Eagle strategy~\citep{Yang2010ES}
\item Egyptian vulture~\citep{Sur2013Egypt}
\item Emperor penguins colony~\citep{Harifi2019}
\item Fish swarm/school~\citep{Li2002Fish}
\item Great salmon run~\citep{Mozaffari2012Salmon}
\item Harris hawks optimization~\citep{Heidari2019}
\item Killer whale algorithm~\citep{Biyanto2017}
\item Krill herd algorithm~\citep{Gandomi2012Krill}
\item Monkey search~\cite{Mucherino2007Monkey}
\end{itemize}

Nature-inspired algorithms have also been developed by drawing inspiration from non-swarm behaviour, physics, chemistry and other biological systems. Examples of such algorithms are
\begin{itemize}
\item Bacterial foraging algorithm~\citep{Passino2002}
\item Big bang-big crunch~\citep{Erol2006}
\item Biogeography-based optimization~\citep{Simon2008BGO}
\item Black hole algorithm~\citep{Hatamlou2012}
\item Charged system search~\citep{Kaveh2010ChSS})
\item Ecology-inspired evolutionary algorithm~\citep{Parpinelli2011eco}
\item Gravitational search~\citep{Rashedi2009GSA}
\item Water cycle algorithm ~\citep{Eskandar2012}
\end{itemize}

It is worth pointing out that some of these algorithms may perform well and provide very competitive results, but other algorithms are not so efficient. The current literature and various studies seem to indicate that their performance and results are quite mixed.

\section{Hybridization}

There are many hybrid algorithms and variants in the current literature. A systematical analysis requires some substantial effort and time to go through all the algorithms and understand their components. However, it is not our intention to do such a complete analysis. Our emphasis here is to outline some of the hybridization schemes that may be relevant to most existing hybrid algorithms and their variants.

\begin{figure}[h]
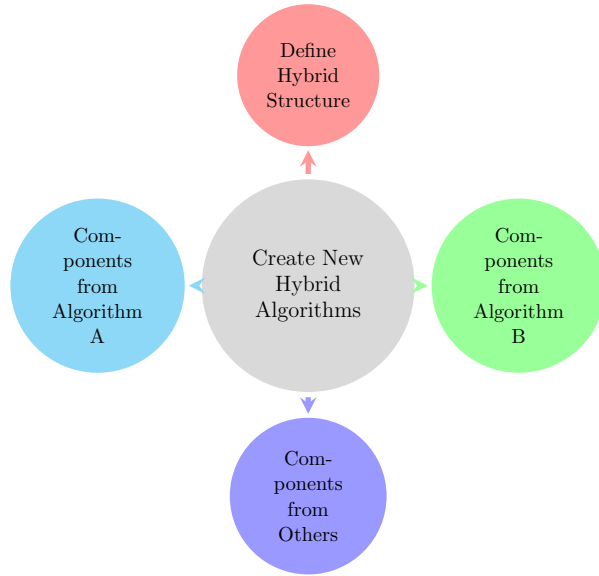
 \begin{center}
\smartdiagramset{planet text width=2.30cm, distance planet-satellite=4cm,
satellite size=2cm,satellite text width=2cm}
\scalebox{0.7}{
\smartdiagram[constellation diagram]{Create New \\ Hybrid \\ Algorithms, Define Hybrid Structure, Components \\ from \\ Algorithm A, Components \\ from \\ Others, Components \\ from \\ Algorithm B} }
\caption{Steps to create a new hybrid algorithm. \label{Hybrid-fig-100}}
\end{center}\end{figure}

Loosely speaking, to create a new hybrid algorithm, researchers tend to draw the good or efficient components from different algorithms.
Imagine that there are two algorithms (A and B) that are reasonably effective, if you want to design a new hybrid algorithm, you may use some components from Algorithm A and some components from Algorithm B to form a new algorithm. However, you have to decide how to put them together nicely, which requires a good structure. In addition, you may also want to use other components such as initialization and randomization from other algorithms and techniques. We can represent this schematically in Fig.~\ref{Hybrid-fig-100}.

Obviously, there are different ways to analyze and classify the hybrid algorithms. One of such studies is to look their purpose and different stages of hybridization by Ting et al.~\citep{Ting2015}. Based on this study, we can now extend it further and schematically summarize hybrid algorithms into four broad schemes.

\subsection{Hybridization Schemes}

It is not an easy task to summarize all the relevant steps and
the actual process for creating a new hybrid algorithm. However, it is possible to give some indications that the potential components and their link structure.  For simplicity and ease in discussion, we now use three different algorithms, namely, Algorithm A, Algorithm B and Algorithm C and others.

\subsubsection{Sequential Hybrid}

\begin{figure}[h]
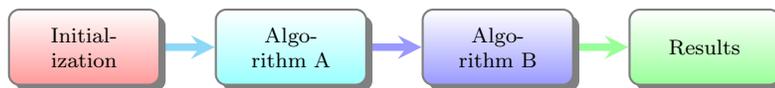

\begin{center}
\smartdiagramset{text width=1.5cm,
font=\scriptsize, back arrow disabled=true}
\smartdiagram[flow diagram:horizontal]{Initialization, Algorithm A, Algorithm B, Results}
\caption{Sequential structure of hybridization. \label{Hybrid-Seq-100} }
\end{center}
\end{figure}

One simple way of designing hybrid algorithms is to use a sequential structure (see Fig.~\ref{Hybrid-Seq-100}). For a given population of $n$ solutions, Algorithm A is run first, then the results are fed into Algorithm B. If needed, another algorithm (say, Algorithm C) is used further.

In practice, both algorithms will be executed iteratively and the final results are processed together.  One additional variation is that the population of $n$ solutions can be split into two or more groups so that each subpopulation is updated by each algorithm.

From numerical simulation and various studies, it seems that this simple structure may be quite popular, but it may not be the best way for hybridization because the solutions are not fully mixed, thus limiting the overall effectiveness of the hybrid algorithm.

\subsubsection{Parallel Hybrid}

\begin{figure}[h]
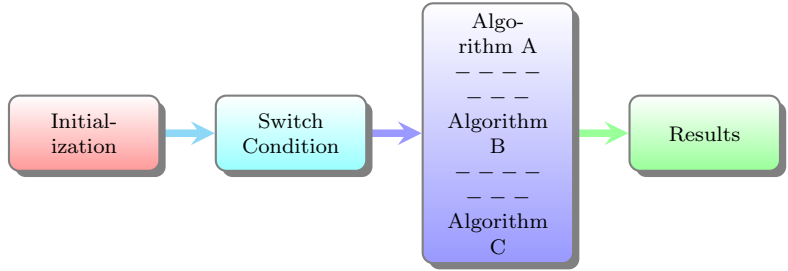

\begin{center}
\smartdiagramset{text width=1.5cm,
font=\scriptsize, back arrow disabled=true}
\smartdiagram[flow diagram:horizontal]{Initialization,
Switch Condition,
Algorithm A \\ $-------$ \\ Algorithm B \\ $-------$ \\ Algorithm C, Results}

\caption{Parallel structure of hybridization. \label{Hybrid-Para-100}}
\end{center}
\end{figure}

Another simple structure for hybridization is to put two or more algorithms in parallel (see Fig.~\ref{Hybrid-Para-100}). There is
often a switch condition, often using a random number, to decide which algorithm to run during each iteration. Another equally popular way is to split the population into subpopulations, and then feed each subpopulation into each algorithm for further iterations. Then, the overall population can be assembled together so as to sort out the best solutions.

Similar to the sequential structure, this structure is also simple and quite popular. However, the solutions may not be fully mixed, thus limiting the diversity of the solutions and consequentially the overall effectiveness of the hybrid.

\subsubsection{Full Hybrid}

In addition to the above simple structures, a more effective way for hybridization is to fully hybridize all the components. In this case, different components from different algorithms are assembled together in the ways like chromosomes as multiple-site crossover. All the components work closely in the whole population, which can often lead to a more effective form of hybridization.

However, the details of each hybridized algorithm may have its own structure, and there is no universal way to achieve a good hybrid.
Care should be taken, because there is no guarantee for any success in the hybridization if all the good components are simply being put together. A pile of good materials does not automatically give a beautiful building, a good architect and multiple engineers are needed to finish the building. Similarly, multiple components of different algorithms do not lead to a good hybrid algorithm. Careful design and extensive numerical tests are required to make it a potentially useful algorithm.

\subsubsection{Mixed Hybrid}

After analyzing various algorithms and their hybrid variants, it seems that many hybrid algorithms have a mixed structure. They can mix the sequential, parallel and full structures into a single algorithm, or they can use some part or aspect of these structures to build a hybrid algorithm. The overall effectiveness of hybrid algorithms can be quite mixed. Some algorithms have some significant improvements and some can only make it work marginally.

Indeed, this is still an open question: How to design a hybrid algorithm effectively? Further research is highly needed in this area.

\subsection{Issues and Warnings}

Despite the extensive research and various studies concerning hybrid algorithms, there are many serious issues that researchers should be aware of. For example, it seems that some variants appeared to be some random combination of some existing algorithms without any careful thinking, and the performance of some hybrid algorithms may be  doubtful.

In the previous writing, we warned the danger of random combinations for hybridization \citep{Ting2015}. Now we highlight this serious issue again here.

Suppose there are $n$ algorithms, if you randomly choose $2 \le k \le n$ algorithms or their components to form (randomly) a so-called new hybrid, then there are
\begin{equation}
C_n^k=\frac{n!}{k! (n-k)!},
\end{equation}
possible combinations. For $n=30$ and $k=2$, there will be 435 hybrids. For $n=30$ and $k=5$, there will be $142 506$ hybrid algorithms.

To demonstrate this serious issue further, let us hypothetically imagine that there are three algorithms: Duck chasing algorithm (DCA), Basil leaf algorithm (BLA), and Star gazing algorithm (SGA).
One should not randomly form absurd algorithms, such as Star-Duck Algorithm, Basil-Star Algorithm, Basil-Leaf-Duck Algorithm, or Star-Basil-Duck Algorithm. No researchers should ever do it (except, perhaps, for a possible chat-bot mutant). In addition, there are millions of plant species and animal species, researchers should not invent millions of algorithms, called apple algorithm, basil algorithm, cucumber algorithm, aardvark algorithm, dodo algorithm, yak algorithm, or zonkey algorithm. New algorithms should be based on true novelties and true efficiency.

Obviously, there are other issues as well. For example, if a hybrid works well, it is not clear how it may work because there are no mathematical or theoretical analysis how these algorithms work in general. In addition, in performance comparison studies, some researchers used the computational time or running time as a measure for comparing different algorithms or variants, but the actual running time on a computer can depend on many factors, including hardware configurations, software used (as well as any potential background anti-virus software), and the implementation details (such as vector-based approach versus a for loop). In this context, there are no universally accepted good performance metrics at the moment. This is still an open problem.

\section{Insights and Recommendations}

Based on the current literature and various studies for analyzing different nature-inspired algorithms \citep{YangBook2014,Yang2020Rev}, we provide some insights into nature-inspired metaheuristic algorithms.

\begin{enumerate}

\item Algorithms can be linear or nonlinear in their solution-update dynamics. For example, PSO is a linear system because its algorithmic equations are linear, but the firefly algorithm is a nonlinear system because Eq.~(\ref{FA-equ-100}) is nonlinear. In general, the characteristics of nonlinear systems tend to be more diverse. For example, the paths traced by individual fireflies can be spare with fractal-like structures, which may explain the search efficiency and good performance of the firefly algorithm.

\item If random walks are used properly, they can improve the search efficiency of an algorithm. For example, L\'evy flights with the step sizes being drawn from a L\'evy distribution tend to have the characteristics of super-diffusion, which can cover a much larger search region than standard diffusive isotropic random walks with steps being drawn from a Gaussian distribution. Cuckoo search uses L\'evy flights, which shows some scale-free properties in the search behaviour.    A few other later algorithms also used L\'evy flights with the intention to improve their performance.

\item Parameter tuning is important for almost all algorithms. Since almost all algorithms have algorithm-dependent parameters, the performance of an algorithm can be influenced by its parameter setting. Thus, the proper tuning of such parameters should be carried out before it can be used to solve optimization problems effectively. However, parameter tuning is itself an optimization problem. Therefore, the tuning of algorithmic parameters can be considered as a hyper-optimization problem because it is the optimization of an optimization algorithm. In fact, how to optimally tune parameter and how to optimally control parameters  during iterations are two open problems.

\item Balance of exploration and exploitation is important, though it is very challenging to achieve it in practice. Theoretically, how to achieve this balance is still an open problem. In practice, some techniques have been used to approximate or estimate this balance. For example, the bat algorithm used the variations of loudness and pulse emission rates to control this balance, whereas the genetic algorithm tends to use a 5:1 rule or 80-20 rule for this. Loosely speaking, about 80\% of the initial search should be about exploration, and about 20\% as exploitation. This can vary according to the iterations in practice.

\item There are many open problems concerning nature-inspired algorithms. For example, there are non unified theoretical framework for analyzing such algorithms mathematically or statistically so as to gain insights into their stability, convergence, rate of convergence and robustness.
    In addition, benchmarking is also an important topic because it is not clear what types of benchmarks are most useful in validating new algorithms. Currently, most benchmarks are smooth functions, which have almost nothing to do with real-world applications.

\end{enumerate}

For the hybrid algorithms, we would like to make the following
recommendations in the future research: Synergy, Structure and Simplicity.

\begin{itemize}
\item \emph{Synergy}: In hybrid algorithms, different components should work together to produce some synergy. Simple use of best components do not necessarily lead to best hybrids or results.
    Obviously, how to achieve a perfect synergy in hybridization is still an open problem.

\item \emph{Structure}: Structure does matter. Since a pile of good-quality building materials does not make it a useful building, a loose assemblage of algorithmic components does not create a good hybrid algorithm. The order, role and strength of each component from different algorithms can be very important. Again, how to achieve this is still an un-resolved issue.

\item \emph{Simplicity}: A simple and clear structure is preferred.
There are multiple ways of putting together different algorithmic components and, if the overall performance is about the same level, then a simpler structure is preferable, not only because it is simpler to implement but also because it may be easier to understand and analyze.

\end{itemize}

We sincerely hope that this principle of synergy, simplicity and structure can inspire further research in this area.  We also hope that more novel and truly effective hybrid algorithms will appear in the future so that more challenging real-world problems can be solved efficiently.

\end{document}